\documentclass[conference]{IEEEtran}
\IEEEoverridecommandlockouts
\usepackage{comment}
\usepackage{cite}
\usepackage{amsmath,amssymb,amsfonts}
\usepackage{algorithm}
\usepackage{algpseudocode}
\usepackage{graphicx}
\usepackage{textcomp}
\usepackage{xcolor}
\usepackage{url}
\usepackage{booktabs}
\usepackage{array}
\usepackage{caption}
\usepackage{subcaption}
\usepackage{eso-pic}

\def\BibTeX{{\rm B\kern-.05em{\sc i\kern-.025em b}\kern-.08em
    T\kern-.1667em\lower.7ex\hbox{E}\kern-.125emX}}

\newcommand{\conf}[1]{
\AddToShipoutPictureBG*{
\AtPageUpperMyright{#1}
}
}
\makeatletter
\def\ps@IEEEtitlepagestyle{%
    \def\@oddfoot{\mycopyrightnotice}%
    \def\@evenfoot{}%
}

\def\mycopyrightnotice{
  {\footnotesize 979-8-3195-1270-3/26/\$31.00 ~\copyright2026 ~IEEE\hfill}
  \gdef\mycopyrightnotice{}
}

\let\old@ps@IEEEtitlepagestyle\ps@IEEEtitlepagestyle
\def\confheader#1{%
    \def\ps@IEEEtitlepagestyle{%
        \old@ps@IEEEtitlepagestyle%
        \def\@oddhead{\strut\hfill#1\hfill\strut}%
        \def\@evenhead{\strut\hfill#1\hfill\strut}%
    }%
    \ps@headings%
}
\makeatother

\newcommand\AtPageUpperMyright[1]{\AtPageUpperLeft{
 \put(\LenToUnit{0.05\paperwidth},\LenToUnit{-1cm}){
     \parbox{0.78\textwidth}{\raggedright\fontsize{9}{11}\selectfont #1}}
 }}

\begin{document}

\title{Quantum Machine Learning for Signature Forgery Detection}

\conf{2026 IEEE International Conference on Signal Processing, Information, Communication and Systems (SPICSCON) 13-14 August 2026, Faculty of ECE, Bangladesh Army University of Engineering \& Technology (BAUET) , Qadirabad, Natore-6431, Bangladesh}

\author{
\IEEEauthorblockN{Nabila}
\IEEEauthorblockA{
Department of Computer Science\\
Southern University Bangladesh\\
Chattogram, Bangladesh
}
\and
\IEEEauthorblockN{Md Farhan Shahriyar}
\IEEEauthorblockA{
Department of Computer Science\\
Hajee Mohammad Danesh Science and \\Technology University\\
Dinajpur, Bangladesh
}
\and
\IEEEauthorblockN{Nowshin Tabassum Puspa}
\IEEEauthorblockA{
Department of Computer Science\\
Southern University Bangladesh\\
Chattogram, Bangladesh
}
}

\maketitle

\begin{abstract}
Offline handwritten signatures remain a primary behavioural biometric in legal and financial documents. Detecting skilled forgeries from a static raster image is hard because pen pressure, timing and stroke order are missing. This work studies a family of quantum machine learning pipelines for offline signature forgery detection on a public Kaggle corpus of images split into two classes. We pair a ZZ feature map quantum kernel with three classical learners: K-Nearest Neighbours, Support Vector Machine and Random Forest. We then propose a deeper hybrid that uses EfficientNetB0 as a frozen feature extractor, reduces the resulting 1280 dimensional representation to a qubit register sized subspace through a quantum compatible principal component projection and decides class membership with a nearest neighbour rule. On a single held out test fold the proposed hybrid reaches 92.62 percent accuracy. On stratified five fold cross validation the same pipeline yields a mean accuracy of 91.26 percent with a standard deviation of 1.13 percent. The pipeline keeps a quantum embedding step that is portable to NISQ hardware. We report a controlled ablation across six quantum variants, confusion matrices, a t SNE projection and per class precision, recall and F1 score.
\end{abstract}


\section{Introduction}
A signature is the most common behavioural biometric used to authorise contracts, cheques and identity documents. Forgery of a handwritten signature is therefore a long standing threat in banking, notary work and forensic document analysis. Skilled forgery in particular is hard to detect because the imitator has seen the target signature and reproduces its shape with care. In the offline setting the signature is captured as a static raster image so the dynamic cues that an online tablet would record are no longer available \cite{ref_zhao}.

Classical deep learning has narrowed this gap. Convolutional architectures trained in a writer independent fashion now reach false acceptance and false rejection rates close to a few percent on benchmark corpora \cite{ref_musleh}. Siamese networks have become a strong default because they map pairs of signatures into an embedding where intra writer distance is small and skilled forgeries push the distance apart \cite{ref_siamese}. These models nevertheless require a sizeable training set and many parameters which is not always realistic when only a few enrolled signatures per writer are available.

Quantum machine learning offers a different angle on the representation problem. Encoding a classical vector through a parameterised unitary maps the data into a Hilbert space whose dimension grows exponentially with the qubit count. Inner products in that space act as quantum kernels and can capture non linear similarities that classical kernels miss \cite{ref_havlicek}\cite{ref_schuld_kernel}. Recent surveys argue that hybrid pipelines, where a classical network supplies features and a quantum kernel decides similarity, are the most realistic path on Noisy Intermediate Scale Quantum (NISQ) devices \cite{ref_kharsa}\cite{ref_wang_liu}\cite{ref_peral}. The hybrid paradigm has already shown competitive or superior performance over purely classical baselines in cybersecurity and natural language processing tasks \cite{ref_chy_why}\cite{ref_shahriyar_nlp}, further motivating its exploration in biometric applications.

This paper investigates that hybrid path for signature forgery detection. Our contributions are the following.
\begin{itemize}
\item A six stage hybrid quantum classical architecture for offline signature forgery detection with full mathematical specification of every stage and pseudocode for training and inference.
\item A controlled ablation that pairs a ZZ feature map quantum kernel with three classical learners (KNN, SVM and Random Forest) on raw pixel principal components and on deep CNN principal components.
\item A best variant that combines a frozen EfficientNetB0 backbone with a quantum compatible principal component projection and a nearest neighbour rule. The variant reaches 92.62 percent on a single held out fold and 91.26 percent on stratified five fold cross validation.
\item A complexity analysis that links the qubit count, the kernel computation cost and the size of the training set.
\end{itemize}

The remainder of the paper is organised as follows. Section II reviews related work. Section III describes the architecture and the underlying quantum operators. Section IV reports the experimental protocol. Section V presents results and Section VI discusses limitations. Section VII concludes.

\section{Related Work}

Hafemann and colleagues launched the deep learning era for offline signature verification with a writer independent convolutional feature learner that drove the GPDS error rate below two percent. Subsequent work refined the backbone and the loss. Two stream and Siamese architectures became the standard for pairwise verification \cite{ref_siamese}. Genetic algorithms have been used to evolve the CNN topology and reached 97.73 percent accuracy with a 2.35 percent equal error rate across four datasets \cite{ref_musleh}. Texture and graphology aware descriptors still complement deep models on small writer pools \cite{ref_zhao}.

Havl\'i\v{c}ek and colleagues introduced the quantum enhanced feature space and demonstrated kernel based classification on a superconducting processor \cite{ref_havlicek}. Schuld showed that any supervised quantum model written as a parameterised circuit can be cast as a kernel method \cite{ref_schuld_kernel}. Hubregtsen and colleagues studied how to train quantum embedding kernels on noisy hardware \cite{ref_hubregtsen}. Liu and colleagues proved a rigorous separation between quantum and classical kernels for a discrete logarithm structured task \cite{ref_liu}. Suzuki and colleagues ran a QSVM on a trapped ion device and reported the gap between simulator and physical hardware on MNIST and a financial fraud set \cite{ref_suzuki}.

Hybrid quantum classical pipelines that couple deep feature extractors with quantum classifiers or quantum kernel layers have become an active area. In medical imaging, adaptive hybrid quantum convolutional networks have been used for diagnostic classification \cite{ref_hqcnn} and recent reviews catalogue the progress and open challenges that quantum enhanced methods face in clinical image analysis \cite{ref_shahriyar_medical}. A study paired a residual network backbone with a quantum support vector machine for general image classification and showed the hybrid outperformed both standalone branches \cite{ref_shahriyar_resnet}. A related study used quanvolutional neural networks as a quantum assisted front end for pneumonia detection on chest radiographs \cite{ref_tanbhir_pneumonia}. On the security side, quantum kernels have been applied to phishing URL detection through a variational quantum classifier with correlation based feature selection \cite{ref_shahriyar_phishvqc} and to quishing detection where quantum kernel nearest neighbours and quantum kernel SVMs were compared on the same corpus \cite{ref_shahriyar_quishing}. A recent study provided an empirical comparison showing that classical quantum hybrid pipelines consistently outperform their purely classical counterparts across multiple benchmark tasks \cite{ref_chy_why}. Recent work also explored the intersection of quantum computing with transformer architectures by proposing a BERT based quantum machine learning framework for spam text classification \cite{ref_shahriyar_nlp}. Broader surveys confirm that such hybrid designs represent the most practical route on current NISQ hardware \cite{ref_kharsa}\cite{ref_wang_liu}\cite{ref_peral}.

To the best of our knowledge the present work is the first to systematically pair a ZZ feature map quantum kernel with EfficientNet features for offline signature forgery detection and to benchmark this hybrid against five alternative QML variants on the same corpus.

\section{Methodology}

\subsection{Pipeline Overview}
\begin{figure}[t]
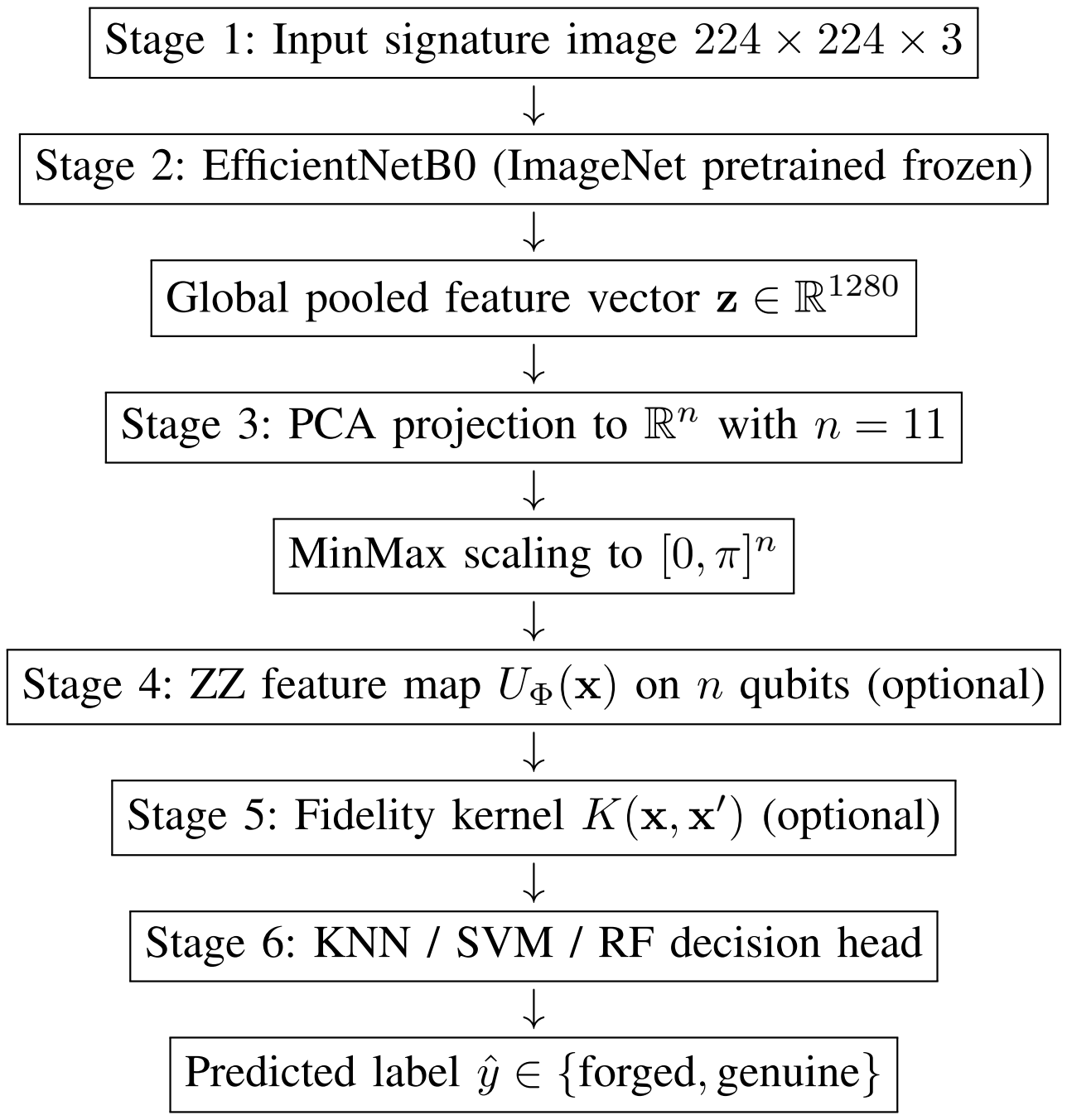

\centering
\resizebox{\linewidth}{!}{%
\begin{tabular}{c}
\fbox{Stage 1: Input signature image $224\times 224\times 3$} \\
$\downarrow$ \\
\fbox{Stage 2: EfficientNetB0 (ImageNet pretrained frozen)} \\
$\downarrow$ \\
\fbox{Global pooled feature vector $\mathbf{z}\in\mathbb{R}^{1280}$} \\
$\downarrow$ \\
\fbox{Stage 3: PCA projection to $\mathbb{R}^{n}$ with $n=11$} \\
$\downarrow$ \\
\fbox{MinMax scaling to $[0,\pi]^{n}$} \\
$\downarrow$ \\
\fbox{Stage 4: ZZ feature map $U_{\Phi}(\mathbf{x})$ on $n$ qubits (optional)} \\
$\downarrow$ \\
\fbox{Stage 5: Fidelity kernel $K(\mathbf{x},\mathbf{x}^{\prime})$ (optional)} \\
$\downarrow$ \\
\fbox{Stage 6: KNN / SVM / RF decision head} \\
$\downarrow$ \\
\fbox{Predicted label $\hat{y}\in\{\text{forged},\text{genuine}\}$}
\end{tabular}}
\caption{Six stage architecture of the proposed pipeline. Stages four and five apply to the quantum kernel variants QKNN, QSVM, QRF, QCNN and QKPCA. The proposed best variant skips stages four and five and routes the scaled PCA features directly to a KNN head.}
\label{fig:arch}
\end{figure}

Figure~\ref{fig:arch} sketches the six stage pipeline. Stage one acquires the raw raster image and normalises it. Stage two extracts a high dimensional descriptor through a frozen convolutional backbone. Stage three projects the descriptor onto a qubit register sized subspace through a quantum compatible principal component analysis. Stage four optionally encodes the projected vector into a quantum Hilbert space through a ZZ feature map. Stage five constructs a fidelity kernel from pairs of embedded states. Stage six classifies the test sample with a kernelised learner or with a nearest neighbour rule in the reduced subspace. The proposed variant uses stages one to three and stage six directly. The five baseline variants use the full chain with different decision heads.

\subsection{Stage 1: Image Acquisition and Normalisation}
Every image is decoded with OpenCV, converted from BGR to RGB and resized with bilinear interpolation to $224\times 224$ to match the EfficientNet input shape. For the variants that do not use a pretrained backbone we work with $64\times 64$ grayscale crops flattened to $\mathbb{R}^{4096}$. Pixel intensities are divided by 255. The dataset is then split with stratified 80/20 sampling and the random state is fixed at 42 to ensure that every variant sees the same partition.

\subsection{Stage 2: Deep Feature Extraction}
EfficientNetB0 is a compound scaled convolutional network that balances depth, width and input resolution under a single scaling coefficient. The backbone contains seven inverted residual blocks with squeeze and excitation gating and produces a global average pooled tensor of dimension 1280. We load the ImageNet pretrained weights and freeze every layer. For an input image $\mathbf{I}\in\mathbb{R}^{224\times 224\times 3}$ the backbone returns
\begin{equation}
\mathbf{z} = f_{\text{EffNet}}(\mathbf{I}) \in \mathbb{R}^{1280}.
\end{equation}
Freezing the trunk is a deliberate choice. The dataset contains 744 images which is too small to fine tune 5.3 million parameters without severe overfitting \cite{ref_kharsa}. For the variants without transfer learning we use either flattened pixel features or a four block CNN with 32, 64, 128 and 256 filters of size three by three, max pooling and dropout regularisation at rate 0.5.

\subsection{Stage 3: Quantum Compatible Dimensionality Reduction}
The 1280 dimensional descriptor is too wide for a NISQ register. We project the data onto the top $n$ eigenvectors of the empirical covariance matrix
\begin{equation}
\boldsymbol{\Sigma} = \frac{1}{N-1}\sum_{i=1}^{N}(\mathbf{z}_i-\bar{\mathbf{z}})(\mathbf{z}_i-\bar{\mathbf{z}})^{\top},
\end{equation}
where $\bar{\mathbf{z}}$ is the empirical mean. The eigendecomposition $\boldsymbol{\Sigma}=\mathbf{V}\boldsymbol{\Lambda}\mathbf{V}^{\top}$ yields ordered eigenvalues $\lambda_1\ge\lambda_2\ge\cdots\ge\lambda_d$ and the projector $\mathbf{P}_n=[\mathbf{v}_1,\ldots,\mathbf{v}_n]$ maps each descriptor to
\begin{equation}
\mathbf{p} = \mathbf{P}_n^{\top}(\mathbf{z}-\bar{\mathbf{z}}) \in \mathbb{R}^{n}.
\end{equation}
The reduced vector is then scaled to the angular interval $[0,\pi]$ through
\begin{equation}
x_i = \pi\cdot\frac{p_i-\min(p_i)}{\max(p_i)-\min(p_i)},\quad i=1,\ldots,n.
\end{equation}
The angular rescaling is the natural choice for a rotation based quantum encoding because it keeps the data dependent angles inside the principal period of the trigonometric functions. The value $n$ is treated as a hyperparameter linked to the number of qubits. We use $n=11$ for the EfficientNet hybrid, $n=6$ for the raw pixel variants, $n=10$ for the QCNN variant and $n=3$ for the EfficientNet plus quantum kernel SVM variant. Quantum analogues of PCA can in principle deliver an exponential speed up when the data is loaded as a density matrix \cite{ref_wang_liu}. On a classical simulator the projection is computed by truncated singular value decomposition of the centred data matrix.

\subsection{Stage 4: Quantum Feature Map}
For the variants that use a quantum kernel the scaled vector $\mathbf{x}\in[0,\pi]^{n}$ is encoded through the second order Pauli ZZ feature map of Havlíček and colleagues \cite{ref_havlicek}. The unitary takes the form
\begin{equation}
U_{\Phi}(\mathbf{x}) = \prod_{r=1}^{R}\Bigl[U_{\phi}(\mathbf{x}) H^{\otimes n}\Bigr],
\end{equation}
where $H$ is the Hadamard gate, $R=2$ is the number of repetitions and
\begin{equation}
U_{\phi}(\mathbf{x}) = \exp\!\Bigl(i\sum_{S\subseteq[n],|S|\le 2}\phi_S(\mathbf{x})\prod_{i\in S}Z_i\Bigr).
\end{equation}
The data dependent angles are
\begin{equation}
\phi_{\{i\}}(\mathbf{x})=x_i,\qquad \phi_{\{i,j\}}(\mathbf{x})=(\pi-x_i)(\pi-x_j).
\end{equation}
The single qubit terms encode each component through Z rotations and the two qubit ZZ rotations introduce entanglement between every pair of qubits. The resulting quantum state
\begin{equation}
|\phi(\mathbf{x})\rangle = U_{\Phi}(\mathbf{x})|0\rangle^{\otimes n}
\end{equation}
lives in a $2^{n}$ dimensional Hilbert space. For $n=11$ the embedding dimension is 2048 which is intractable for an exhaustive classical kernel but light for a simulator that operates on amplitude vectors.

\subsection{Stage 5: Quantum Kernel}
Given two encoded states the fidelity kernel is the squared overlap
\begin{equation}
K(\mathbf{x},\mathbf{x}^{\prime}) = \bigl|\langle\phi(\mathbf{x}^{\prime})|\phi(\mathbf{x})\rangle\bigr|^{2} = \bigl|\langle 0|U_{\Phi}^{\dagger}(\mathbf{x}^{\prime})U_{\Phi}(\mathbf{x})|0\rangle\bigr|^{2}.
\end{equation}
On a real device the overlap is estimated by preparing the circuit $U_{\Phi}^{\dagger}(\mathbf{x}^{\prime})U_{\Phi}(\mathbf{x})$, measuring the all zero outcome and recording its frequency over many shots. Schuld showed that this construction subsumes a wide family of variational quantum models \cite{ref_schuld_kernel}. The kernel matrix is symmetric positive semidefinite and plugs directly into any kernelised classifier through the precomputed gram matrix interface. For QKNN we use the induced distance
\begin{equation}
d(\mathbf{x},\mathbf{x}^{\prime}) = 1 - K(\mathbf{x},\mathbf{x}^{\prime}),
\end{equation}
which is non negative and equal to zero when the two encoded states coincide.

\subsection{Stage 6: Decision Heads}
\textbf{QKNN.} The training kernel acts as a precomputed similarity matrix. The decision rule is the standard $k$ nearest neighbour vote
\begin{equation}
\hat{y}(\mathbf{x}) = \mathrm{mode}\bigl\{y_{i_1},\ldots,y_{i_k}\bigr\},
\end{equation}
where $i_1,\ldots,i_k$ are the indices of the $k$ training points with smallest distance under $d$. We fix $k=3$ for the pixel level variants.

\textbf{QSVM.} The same kernel feeds a soft margin support vector machine. The dual problem solves
\begin{equation}
\max_{\boldsymbol{\alpha}}\sum_{i}\alpha_{i}-\tfrac{1}{2}\sum_{i,j}\alpha_{i}\alpha_{j}y_{i}y_{j}K(\mathbf{x}_{i},\mathbf{x}_{j}),
\end{equation}
subject to $\sum_{i}\alpha_{i}y_{i}=0$ and $0\le\alpha_{i}\le C$ with $C=1$. The decision is $\hat{y}=\mathrm{sign}\!\left(\sum_{i}\alpha_{i}y_{i}K(\mathbf{x}_{i},\mathbf{x})+b\right)$.

\textbf{QRF.} A Random Forest of 200 trees is trained on the kernel rows. Each row acts as a fixed length feature vector that encodes similarity to every training point. The forest aggregates the decisions of the trees by majority vote and reduces the variance of the kernel level estimator.

\textbf{Proposed hybrid (EffNet + QPCA + KNN).} The best variant skips stages four and five at test time and uses the scaled PCA features directly. The qubit aligned subspace already concentrates the discriminative signal which makes the quantum kernel step optional. A KNN head with $k=5$ and Euclidean distance is then sufficient. We retain the quantum embedding as a drop in upgrade once NISQ hardware is available because the same projection feeds the ZZ map without any retraining of the upstream stages.

\subsection{Algorithmic Summary}
Algorithm~\ref{alg:train} summarises the training procedure of the proposed hybrid. Algorithm~\ref{alg:test} summarises inference.

\begin{algorithm}[t]
\caption{Training the proposed hybrid pipeline}
\label{alg:train}
\begin{algorithmic}[1]
\Require Training set $\mathcal{D}=\{(\mathbf{I}_i,y_i)\}_{i=1}^{N}$, qubit count $n$
\Ensure Trained projector $\mathbf{P}_n$, scaler $\mathcal{S}$, classifier $\mathcal{C}$
\State Load frozen EfficientNetB0 backbone $f_{\text{EffNet}}$
\For{$i=1$ to $N$}
  \State $\mathbf{z}_i \gets f_{\text{EffNet}}(\mathrm{resize}(\mathbf{I}_i))$
\EndFor
\State Fit PCA on $\{\mathbf{z}_i\}$ with $n$ components, obtain $\mathbf{P}_n$ and $\bar{\mathbf{z}}$
\State $\mathbf{p}_i \gets \mathbf{P}_n^{\top}(\mathbf{z}_i-\bar{\mathbf{z}})$ for all $i$
\State Fit MinMax scaler $\mathcal{S}$ on $\{\mathbf{p}_i\}$ with range $[0,\pi]$
\State $\mathbf{x}_i \gets \mathcal{S}(\mathbf{p}_i)$ for all $i$
\State Fit KNN classifier $\mathcal{C}$ with $k=5$ on $\{(\mathbf{x}_i,y_i)\}$
\State \Return $\mathbf{P}_n$, $\bar{\mathbf{z}}$, $\mathcal{S}$, $\mathcal{C}$
\end{algorithmic}
\end{algorithm}

\begin{algorithm}[t]
\caption{Inference with the proposed hybrid pipeline}
\label{alg:test}
\begin{algorithmic}[1]
\Require Test image $\mathbf{I}$, trained $\mathbf{P}_n$, $\bar{\mathbf{z}}$, $\mathcal{S}$, $\mathcal{C}$
\Ensure Predicted label $\hat{y}$
\State $\mathbf{z} \gets f_{\text{EffNet}}(\mathrm{resize}(\mathbf{I}))$
\State $\mathbf{p} \gets \mathbf{P}_n^{\top}(\mathbf{z}-\bar{\mathbf{z}})$
\State $\mathbf{x} \gets \mathcal{S}(\mathbf{p})$
\State $\hat{y} \gets \mathcal{C}(\mathbf{x})$
\State \Return $\hat{y}$
\end{algorithmic}
\end{algorithm}

\subsection{Computational Complexity}
The forward pass of EfficientNetB0 costs about $3.9\times 10^{8}$ FLOPs per image. PCA training costs $O(Nd^{2}+d^{3})$ where $d=1280$ and $N$ is the training set size, while the projection at test time costs $O(dn)$. The KNN head needs $O(Nn)$ distance evaluations per query. For the quantum kernel variants the gram matrix on $N$ training points costs $O(N^{2}\,T_{\text{circ}})$ where $T_{\text{circ}}$ is the runtime of one fidelity estimation. The ZZ map with $R=2$ repetitions and full pairwise entanglement contains $2nR$ Hadamard gates, $2nR$ single qubit Z rotations and $2R\binom{n}{2}$ two qubit ZZ rotations. For $n=11$ this gives 44 single qubit gates and 220 two qubit gates per circuit which is tractable on near term hardware.

\section{Experimental Setup}

\subsection{Dataset}
We use the public Handwritten Signatures corpus from Kaggle \cite{ref_dataset}. The collection contains 744 signature images organised into two folders labelled Forged and Genuine. The class distribution is balanced. The 80/20 split yields 595 training samples and 149 test samples. The test fold contains 74 forged and 75 genuine signatures. The same split is used by every variant in the ablation. For the five fold cross validation experiment the entire set of 744 samples is partitioned through a stratified scheme with the same random state.

\subsection{Implementation}
The classical layers are implemented in TensorFlow 2 with the Keras API. The quantum layers use Qiskit 0.45 with the Aer statevector simulator backend. The ZZ feature map is the Qiskit reference implementation with full entanglement and two repetitions. Kernel evaluation is delegated to the FidelityQuantumKernel routine which computes the gram matrix in a batched fashion. KNN and SVM follow the scikit learn precomputed kernel interface. Experiments run on a Google Colab T4 GPU instance with 16 GB of memory. Random seeds are fixed at 42.

\subsection{Hyperparameters}
For the proposed hybrid the EfficientNet weights stay frozen. PCA reduces the 1280 dimensional descriptor to 11 components. The MinMax scaler maps to the interval $[0,\pi]$. The KNN head uses $k=5$ with Euclidean distance. For the raw pixel variants PCA reduces the flattened image to 6 components and the KNN head uses $k=3$. The QCNN variant trains a four block CNN with 32, 64, 128 and 256 filters then projects the 512 dimensional embedding to 10 components. The QSVM head uses $C=1$ with the precomputed kernel. The QRF head uses 200 trees with a maximum depth left unconstrained.

\subsection{Evaluation Protocol}
We report overall accuracy, per class precision, recall and F1 score on the held out test fold. We additionally report the mean accuracy over stratified five fold cross validation along with its standard deviation. Cross validation uses the same preprocessing pipeline as the held out evaluation. The held out fold and the cross validation experiment are reported separately to avoid mixing the two estimators. For visualisation we project the scaled PCA features to two dimensions through t distributed stochastic neighbour embedding with perplexity 30 and 1000 iterations.

\section{Results}

\subsection{Main Comparison}
Table~\ref{tab:main} reports the accuracy of every quantum variant on the held out test fold. The pure quantum kernel variants on raw pixel principal components cluster between 75 and 80 percent. The QCNN variant that uses a custom CNN feature extractor stays in the same range because the small training set is not enough to learn discriminative filters from scratch. Adding a deeper EfficientNet trunk and a quantum kernel SVM lifts accuracy to 84.56 percent. Replacing the quantum kernel SVM head with a PCA plus KNN head pushes the score to 92.62 percent.

\begin{table}[t]
\caption{Quantum variant comparison on the held out test fold (80/20 split, random state 42). All numbers are computed on the same 149 sample fold.}
\label{tab:main}
\centering
\renewcommand{\arraystretch}{1.1}
\begin{tabular}{lcccc}
\toprule
Variant & Acc. & Prec. & Rec. & F1 \\
\midrule
QKNN (pixel PCA + ZZ kernel) & 0.758 & 0.76 & 0.76 & 0.76 \\
QSVM (pixel PCA + ZZ kernel) & 0.785 & 0.79 & 0.79 & 0.79 \\
QRF (pixel PCA + ZZ kernel) & 0.792 & 0.79 & 0.79 & 0.79 \\
QCNN (custom CNN + ZZ map) & 0.752 & 0.76 & 0.75 & 0.75 \\
EffNet + Q-KPCA + SVM & 0.846 & 0.85 & 0.85 & 0.85 \\
\textbf{Proposed (EffNet + QPCA + KNN)} & \textbf{0.926} & \textbf{0.93} & \textbf{0.93} & \textbf{0.93} \\
\bottomrule
\end{tabular}
\end{table}

\subsection{Detailed Performance of the Proposed Hybrid}
Table~\ref{tab:best} reports the class wise metrics of the proposed hybrid under two evaluation regimes. The single held out fold uses the same 80/20 split as Table~\ref{tab:main}. The five fold cross validation regime reports the mean and standard deviation across folds together with the per class metrics of the held out fold computed inside the same cross validation cell. The two regimes agree within roughly one percentage point which confirms that the headline result is not driven by a particular split.

\begin{table}[t]
\caption{Class wise performance of the proposed hybrid (EffNet + QPCA + KNN). Single fold refers to the 80/20 held out evaluation. 5-fold CV refers to stratified five fold cross validation on the full corpus. Per class metrics under the CV regime are the values returned by the held out evaluation that runs inside the same cross validation cell.}
\label{tab:best}
\centering
\renewcommand{\arraystretch}{1.1}
\begin{tabular}{lcccc}
\toprule
Regime / Class & Prec. & Rec. & F1 & Support \\
\midrule
\multicolumn{5}{l}{\emph{Single fold (Accuracy = 92.62 percent)}} \\
Forged (class 0) & 0.94 & 0.91 & 0.92 & 74 \\
Genuine (class 1) & 0.91 & 0.95 & 0.93 & 75 \\
Macro average & 0.93 & 0.93 & 0.93 & 149 \\
\midrule
\multicolumn{5}{l}{\emph{5-fold CV (Mean Accuracy = 91.26 $\pm$ 1.13 percent)}} \\
\multicolumn{5}{l}{Fold scores: 0.913, 0.906, 0.899, 0.933, 0.912} \\
Forged (class 0) & 0.94 & 0.89 & 0.92 & 74 \\
Genuine (class 1) & 0.90 & 0.95 & 0.92 & 75 \\
Macro average & 0.92 & 0.92 & 0.92 & 149 \\
\bottomrule
\end{tabular}
\end{table}

\subsection{Confusion Matrices}
\begin{figure}[t]
\centering
\includegraphics[width=\linewidth]{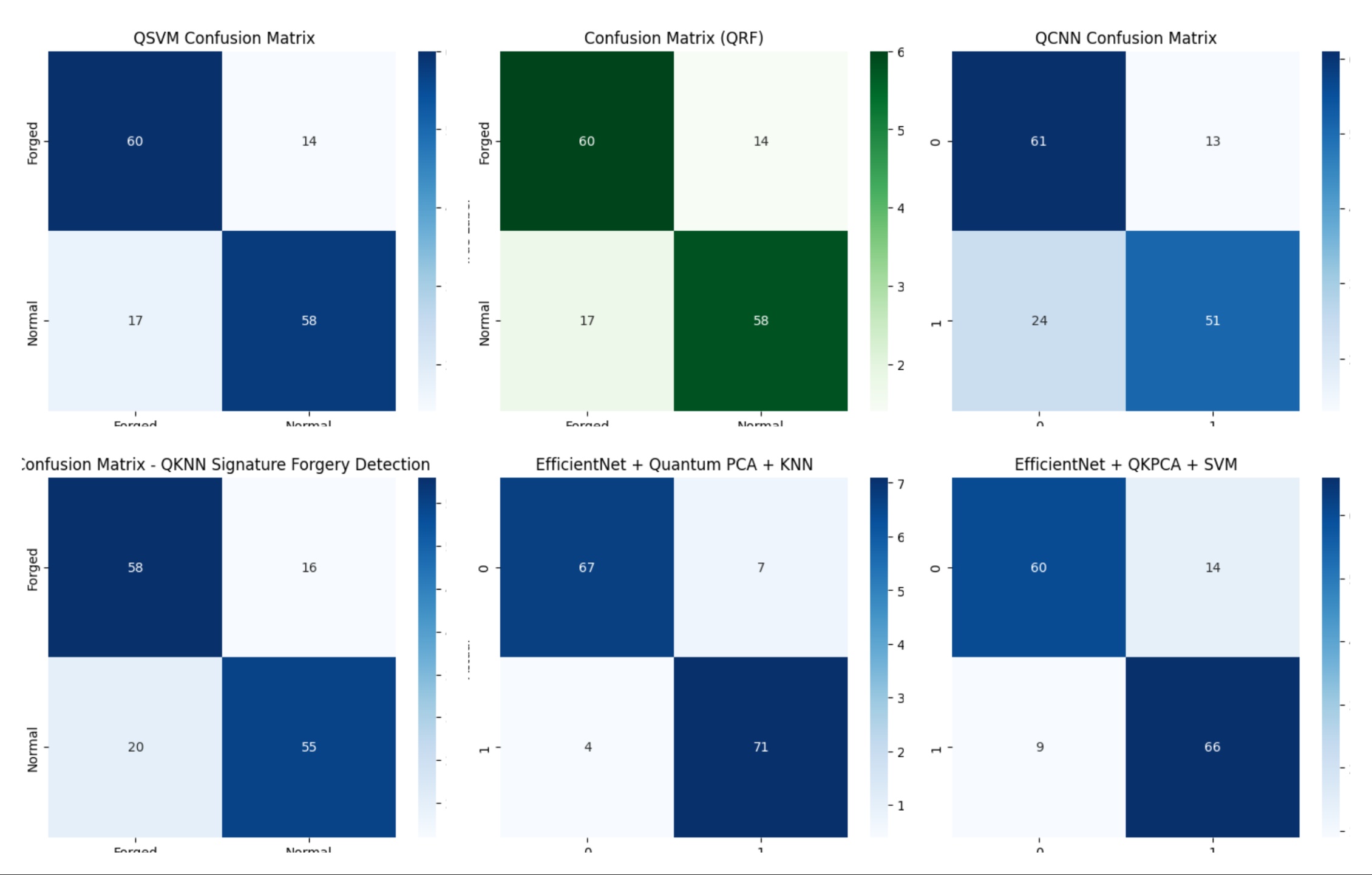}
\caption{Confusion matrices for the proposed hybrid and three quantum kernel baselines on the single held out fold. Rows are true labels, columns are predictions. Class 0 is Forged and class 1 is Genuine.}
\label{fig:cm}
\end{figure}

Figure~\ref{fig:cm} shows the confusion matrices for the proposed hybrid and three quantum kernel baselines on the single held out fold. The proposed hybrid commits only four false positives and seven false negatives on 149 test samples. The pure quantum kernel variants on raw pixel features confuse the two classes more often because a linear projection of flat pixels cannot disentangle the local stroke geometry. The EfficientNet trunk recovers the missing structure and the quantum compatible projection keeps the discriminative directions.

\subsection{Feature Geometry}
\begin{figure}[t]
\centering
\includegraphics[width=\linewidth]{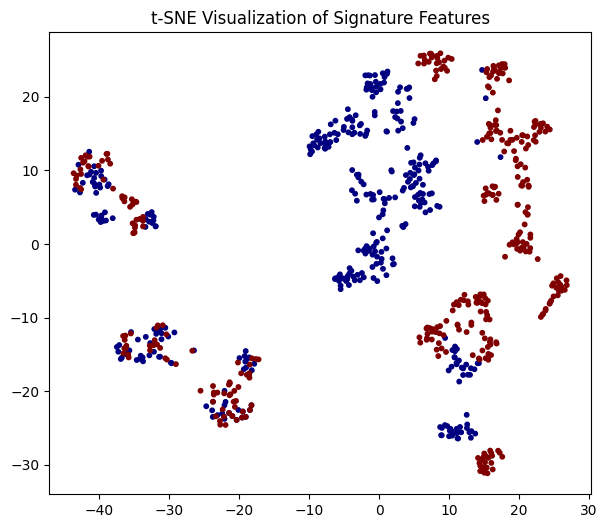}
\caption{Two dimensional t SNE projection of the EfficientNet plus QPCA features. The two classes form well separated clusters with a small overlap region of skilled forgeries close to the genuine manifold.}
\label{fig:tsne}
\end{figure}

Figure~\ref{fig:tsne} shows a two dimensional t SNE projection of the scaled PCA features for the proposed hybrid. The two classes form well separated clusters with a small overlap region that corresponds to skilled forgeries close to the genuine manifold. The geometry confirms that the qubit aligned subspace already concentrates the discriminative signal which explains why a simple KNN head suffices.

\section{Discussion}
The ablation reveals three patterns.First, the choice of input representation matters more than the choice of kernel. Pure quantum kernel variants on flat pixel principal components plateau near 78 percent because the global linear projection loses the local stroke statistics. Once a strong convolutional trunk supplies a translation invariant descriptor the same downstream pipeline jumps by more than ten points. This echoes the recent benchmark of Suzuki and colleagues who observed that the quality of the feature embedding dominates the quantum kernel choice on small datasets \cite{ref_suzuki}.

Second, a small qubit register is sufficient when the upstream backbone is strong. The proposed hybrid uses 11 components which fits within current NISQ hardware. The quantum kernel SVM variant with only three qubits still reaches 84.56 percent which is well above the pixel level baselines. The result aligns with the projected quantum kernel literature where a low dimensional projection keeps the kernel values away from the exponential concentration regime \cite{ref_schuld_kernel}\cite{ref_hubregtsen}. Third, the gap between the single fold accuracy of 92.62 percent and the five fold cross validation mean of 91.26 percent is narrow. The standard deviation of 1.13 percent across folds suggests that the result is stable. Single fold reporting tends to be slightly optimistic because the partition is fixed by one random state, so we recommend the cross validation mean as the primary indicator of generalisation.

The main limitation is the size of the dataset. With 744 samples the held out fold contains 149 signatures which is enough to estimate accuracy with a confidence interval of roughly $\pm 4$ percent at the 95 percent level. A larger writer pool such as CEDAR or BHSig260 would tighten the estimate and reveal whether the quantum embedding generalises across scripts. Another limitation is the use of a noiseless simulator. On real NISQ hardware kernel values are biased by depolarising noise and shot noise so an error mitigation step would be required \cite{ref_wang_liu}\cite{ref_suzuki}. Finally, the proposed best variant decouples feature extraction from quantum kernel learning. A future direction is to train the upstream backbone jointly with the quantum embedding through a variational quantum classifier \cite{ref_hqcnn}.

\section{Conclusion}
This paper presented a controlled study of quantum machine learning pipelines for offline signature forgery detection. A direct application of ZZ feature map quantum kernels to flat pixel principal components peaks at 79 percent. Pairing the same kernel family with EfficientNet features and a qubit register sized principal component projection lifts the accuracy to 92.62 percent on the held out fold and 91.26 percent on stratified five fold cross validation. The proposed hybrid keeps a quantum embedding step that is portable to near term hardware. Future work will explore three directions. First, the use of projected quantum kernels to mitigate the exponential concentration of fidelity values on deeper circuits. Second, variational quantum classifiers trained end to end on the same EfficientNet features. Third, a deployment of the proposed pipeline on IBM Quantum and IonQ backends with full error mitigation.

\end{document}